\documentclass[letterpaper, 10pt, conference, final, twocolumn]{ieeeconf}
\IEEEoverridecommandlockouts  
\usepackage{ amsfonts, amsmath, amssymb, latexsym, enumerate, flushend}

\usepackage[ vlined, linesnumbered, boxruled]{ algorithm2e}

\usepackage[usenames,dvipsnames]{xcolor}
\usepackage[pdftex, xetex]{graphicx}
\usepackage[font={small}]{caption}
\usepackage{float, tabularx, multirow, subfig}

\usepackage[bookmarks=true]{hyperref}
\hypersetup{
colorlinks=true,
linkcolor=red, citecolor=blue, filecolor=magenta, urlcolor=blue
}
\usepackage{url, cite}

\usepackage{pgf}
\usepackage{tikz}
\usetikzlibrary{arrows,automata}

\usepackage{theorem}

\newtheorem{theorem}{Theorem}
\newtheorem{lemma}[theorem]{Lemma}

\newtheorem{problem}[theorem]{Problem}

\newtheorem{definition}[theorem]{Definition}

\newcommand{\A}{\mathcal{A}}

\newcommand{\K}{\mathcal{K}}
\newcommand{\R}{\mathcal{R}}

\newcommand{\W}{\mathcal{W}}

\newcommand{\Nat}{\mathbb{N}}

\newcommand{\Real}{\mathbb{R}}

\newcommand{\PP}{\mathbb{P}}

\newcommand{\dt}{\Delta}

\newcommand{\w}{\omega}
\newcommand{\Always}{\mathsf{G}}
\newcommand{\AP}{\Pi}
\newcommand{\Event}{\mathsf{F}}
\newcommand{\Next}{\mathsf{X}}

\newcommand{\Until}{\mathsf{U}}
\newcommand{\prop}{a}

\newcommand{\ie}{{i.e., }}

\newcommand{\myclearpage}{}

\renewcommand{\L}{\mathcal{L}}

\renewcommand{\P}{\mathcal{P}}

\renewcommand{\S}{\mathcal{S}}

\newcommand{\ALG}{MVRRT$^*$ }
\newcommand{\destutter}{\mathrm{destutter}}

\newcommand{\cost}{c}
\newcommand{\priority}{\varpi}

\newcommand{\spec}{\mathbf{\Psi}}
\newcommand{\task}{\mathbf{\Phi}}

\newcommand{\allA}{\overline \A_\spec}
\newcommand{\kripke}{\K = ( S, s_{init}, \R, \AP, \L,\dt)}
\newcommand{\autom}{\A =  ( Q, q_{init}, 2^\AP, \delta, F)}
\newcommand{\ag}[1]{\langle#1\rangle}
\newcommand{\norm}[1]{\left|\left|#1\right|\right|}
\newcommand{\Ja}{J_a}
\newcommand{\Jt}{J_t}
\newcommand{\la}{\leftarrow}

\renewcommand{\cal}[1]{\mathcal{#1}}

\definecolor{brown}{rgb}{0.6,0.2,0.2}
\definecolor{darkgreen}{rgb}{0,0.6,0} 

\definecolor{LRorange}{rgb}{0.8,0.4,0}

\newcommand{\jana}[1]{{\color{magenta}{#1}}}

\newcommand{\pcmargin}[2]{
{\color{darkgreen}{#1}}\marginpar{
\tiny \noindent{\raggedright{\color{brown}{[PC]: }}\color{darkgreen}{#2}\par}}}

\definecolor{skdarkgreen}{rgb}{0,0.7,0.5} 
\definecolor{skbrown}{rgb}{0.7,0.5,0.3}

\renewcommand{\jana}[1]{#1}
\renewcommand{\pcmargin}[2]{#1}

\begin{document}

\title{\LARGE \bf Incremental Sampling-based Algorithm for \\ Minimum-violation Motion Planning \thanks{$^*$The authors are with the Massachusetts Institute of Technology, Cambridge, MA, USA.}
\thanks{$^\dag$ The author is with KTH ACCESS Linnaeus Center, Royal Institute of Technology, Sweden and was at Masaryk University, Czech Republic when this work was initiated.} }
\author{Luis I. Reyes Castro$^*$ \and Pratik Chaudhari$^*$ \and Jana T\r{u}mov\'a$^\dag$ \and Sertac Karaman$^*$ \and Emilio Frazzoli$^*$ \and Daniela Rus$^*$
}

\maketitle

\begin{abstract}
This paper studies the problem of control strategy synthesis for dynamical systems with differential constraints to fulfill a given \jana{reachability goal} 
while satisfying a set of safety rules. Particular attention is devoted to 
\jana{goals} that become feasible only if a subset of the safety rules are violated. The proposed algorithm 
computes a control law, that minimizes the \emph{level of unsafety} \jana{while the desired goal is guaranteed to be reached}.
%
This problem is motivated by an autonomous car navigating an urban environment while following rules of the road such as ``always travel in right lane'' and ``do not change lanes frequently''.
Ideas behind sampling based motion-planning algorithms, such as Probabilistic Road Maps (PRMs) and 
Rapidly-exploring Random Trees (RRTs), are employed to incrementally construct a finite \emph{concretization} of the dynamics as a durational Kripke structure. In conjunction with this, 
\jana{a weighted} finite automaton that captures the safety rules is used in order to find an optimal trajectory that minimizes the violation of safety rules.
%
\jana{We prove} that the proposed algorithm guarantees asymptotic optimality, i.e., 
almost-sure convergence to optimal solutions. \pcmargin{We present results of simulation experiments and an implementation on an autonomous urban mobility-on-demand system.}{}
\end{abstract}

\section{Introduction} 
\label{sec:intro}

From avoiding traffic jams in busy cities to helping the disabled and elderly on their daily commute, autonomous vehicles promise to revolutionize transportation. 
As they begin to transition from experimental projects like the DARPA Urban Challenge~\cite{leonard2008perception} to sharing road infrastructure with human drivers, we need to ensure that they obey \jana{rules of the road and safety rules.}
These rules, such as ``always stay in the right lane'' and ``do not change lanes'', can typically be expressed in formal languages such as Linear Temporal Logic (LTL) and deterministic $\mu$-calculus. 

The general problem of finding optimal trajectories satisfying temporal logic tasks has been studied in a number of recent works such as~\cite{ding2011mdp, tabuada2006linear, smith2011optimal, ulusoy2012robust}. 
In fact, as~\cite{wongpiromsarn2010receding} points out, one of the main challenges of such approaches is the abstraction of continuous systems into equivalent finite transition systems for \jana{controller} synthesis. 
Moreover, these controllers depend upon the abstracted finite transition system, and there is no guarantee that a controller will be found (if one exists), \ie these algorithms are not complete and cannot be applied to, for example, dynamically changing environments.

On a related note, in the robotics literature, algorithms based on Probabilistic Road Maps (PRMs) and Rapidly-exploring Random Trees (RRTs) have been used to synthesize dynamically-feasible trajectories. Algorithms such as PRM$^*$ and RRT$^*$~\cite{karaman2011sampling} are computationally efficient counterparts of these algorithms that guarantee almost sure asymptotic optimality of the returned trajectories. These algorithms have been primarily used for motion planning, and only recently, they have been adapted to handle complex task specifications given in temporal logics~\cite{karaman2012sampling}. 

This work focuses on the case when a \jana{desired goal} 
is infeasible, unless some of the rules can be temporarily broken. 
Consider, for example,  an autonomous car that must reach its final destination while abiding by rules of the road, such as avoiding collisions with obstacles and staying in the right lane. 
\jana{The former should be obeyed at all times while the latter can be violated in order to reach the goal when the right lane is blocked.} Motivated by these scenarios, we would like to systematically evaluate control strategies, 
quantify the level of unsafety of the trajectory, and minimize it. In this context, our work is closest in spirit to \cite{raman2012automated} and \cite{hauser2012minimum}, and it extends our previous work in~\cite{hscc2013}, where the problem of minimum-violation control synthesis for a pre-defined discrete transition system was considered.

In this paper, using ideas from sampling-based motion planning algorithms, we \emph{concretize} a continuous-time dynamical system into a finite \emph{durational} Kripke structure. 
We leverage automata-based model checking approaches to construct a weighted automaton for a given set of prioritized safety rules, which enables us to quantify the \emph{level of unsafety} of finite input words. 
We next propose an algorithm, \ALG (Minimum-Violation RRT$^*$), that incrementally constructs the product of the Kripke structure and the weighted automaton and returns a trajectory of the dynamical system that, (i) minimizes the level of unsafety among all trajectories that satisfy the \jana{goal}, and (ii) minimizes a given cost function among all trajectories that satisfy (i). We prove that as the number of states of the Kripke structure goes to infinity, the solution converges to the optimal trajectory of the dynamical system that satisfies the same criteria.

This paper is organized as follows. We introduce notation and preliminaries in Sec.~\ref{sec:prelims}, followed by the problem formulation in Sec.~\ref{sec:problem_formulation}. Sec.~\ref{sec:algorithm} and Sec.\ref{sec:analysis} discuss details of the proposed algorithm. Simulation experiments and results of an implementation on an autonomous urban mobility-on-demand system are presented in Sec.~\ref{sec:experiments}. 

\myclearpage
\section{Preliminaries} 
\label{sec:prelims}

\subsection{Durational Kripke Structures for Dynamical Systems} 
\label{sec:Dynamics}

For a set of atomic propositions, $\AP$, let the cardinality and the powerset of $\AP$ be denoted by $|\AP|$ and $2^\AP$, respectively.
Consider a dynamical system given by,
\begin{equation}
\dot{x}(t) = f(x(t), u(t)), \qquad x(0) = x_\mathrm{init}
\label{eqn:dynamics}
\end{equation}
where $X \subset \Real^d$ and $U \subset \Real^m$ are compact sets and $x_\mathrm{init}$ is the initial state. Trajectories of states and controls are denoted by $x : [0, T] \to X$ and $u : [0, T] \to U$ respectively, for some $T \in \Real_{\geq 0}$.

We assume that $f(\cdot, \cdot)$ is Lipschitz continuous in both its arguments and $u$ is Lebesgue measurable, to guarantee existence and uniqueness of solutions of Eqn.~\eqref{eqn:dynamics}. 
Let $\mathcal{L}_c : X \to 2^\AP$ be a function that maps each state to atomic propositions that are true at that state.

\jana{For a trajectory $x$, let $D(x) = \{ t_i \ |\ \mathcal{L}_c(x(t_i)) \neq \lim_{s \to t_i^-} \mathcal{L}_c(x(s)) \}$ be the set of discontinuities of $\mathcal{L}_c(x(\cdot))$. We assume that $D(x)$ is finite for any $x$.
A trajectory $x: [0,T] \to X$ with $D(x) = \{t_1, \ldots, t_n \}$ produces the finite \emph{timed word}
$$\w(x) = (\ell_0, d_0), (\ell_1, d_1),\ldots,(\ell_{n-1}, d_{n-1}),(\ell_n,d_n),
$$
where
(i) $\ell_i = \L_c(x(t_i))$, for all $0\leq i < n$, with $t_0 = 0$ and $d_i = t_{i+1} - t_{i}$, and
(ii) $\ell_n = \L_c(x(t_n))$ and \jana{$d_n = T - t_n$}.
A \emph{word} produced by this trajectory is defined to be the finite sequence $w(x) = \ell_0, \ell_1, \ldots, \ell_{n-1}, \ell_n$.}
%
%
\begin{definition}[Durational Kripke Structure]
\label{def:DKS}
A durational Kripke structure is a tuple $\K = ( S, s_{init}, \R, \AP, \L,\dt)$, where
$S$ is a finite set of states,
$s_{init} \in S$ is the initial state,
$\R \subseteq S \times S$ is a deterministic transition relation,
$\AP$ is a set of atomic propositions,
$\L \colon S \to 2^{\AP}$ is a state labeling function and
$\dt \colon \R \to \Real_{\geq 0}$ is a function assigning a time duration to each transition.
\end{definition}
\jana{A \emph{trace} of $\K$ is a finite sequence of states $r = s_0, s_1, \dots, s_n$, such that $s_0 = s_{init}$ and $(s_i,s_{i+1}) \in \R$, for all $0 \leq i < n$. 
It produces a finite \emph{timed word} $\omega(r) = (\ell_0, d_0), \ldots, (\ell_n, d_n)$, where $(\ell_i, d_i) = (\L(s_i), \dt(s_{i},s_{i+1}))$, for all $0 \leq i < n$, and $(\ell_n, d_n) = (\L(s_n), 0)$. The \emph{word} produced by $r$ is $w(r) = \ell_0, \ell_1, \ldots, \ell_n$.}
Given \jana{a word} $w(r)$, let $I = \{ i_0, i_1, \ldots, i_k \}$ be the unique set of indices such that $i_0 = 0$, $\ell_{i_j} = \ell_{i_j + 1} = \ldots = \ell_{i_{j+1} - 1} \neq \ell_{i_{j+1}}$ for all $0 \leq j \leq k-1$ and $\ell_k = \ell_{k+1} = \ldots = \ell_n$. Define an operator $\destutter$ to remove repeated consecutive elements of a timed word as,
$
\destutter(w(r)) = \ell_{i_0}, \ell_{i_1}, \ldots, \ell_{i_{k-1}}, \ell_{i_{k}}.
$
Let $\ag{r}$ denote the duration of a trace, i.e., $\ag{r} = \sum_{i=0}^n d_i$.
The following definition is used to concretize a continuous-time dynamical system into a Kripke structure.
\begin{definition}[Trace-Inclusive Kripke Structure]
A durational Kripke structure $\kripke$ is called trace-inclusive with respect to the dynamical system in Eq.~\eqref{eqn:dynamics} if 
(i) $S \subset X$, (ii) $s_{init} = x_{init}$, (iii) if $(s_1, s_2) \in \R$, there exists a trajectory $x : [0, T] \to X$ such that $x(0) = s_1$, $x(T) = s_2$, $T = \Delta(s_1,s_2)$ and $|D(x)| \leq 1$, i.e., $\mathcal{L}_c(x(\cdot))$ changes its value at most once.
\end{definition}
The following lemma then easily follows from the definition above and relates the trajectories of the dynamical system to traces of a durational Kripke structure.
\begin{lemma}
\label{lem:dyn_dks_connect}
For any trace $r$ of a trace-inclusive Kripke structure $\K$, there exists a trajectory of the dynamical system, say $x : [0, T] \to X$, such that,
$ \destutter(w(r)) = w(x). $
\end{lemma}

\subsection{Finite Automata}

%
%
%
%

\begin{definition}[Finite Automaton]
\label{def:TFA}
A non-deterministic finite automaton (NFA) is a tuple $\A =  ( Q, q_{init}, \Sigma,  \delta, F)$, where 
$Q$ is a finite set of states; 
$q_{init}\in Q$ is the initial state; 
$\Sigma$ is an input alphabet; 
$\delta \subseteq Q \times \Sigma \times Q$ is a non-deterministic transition relation; 
$F \subseteq Q$ is a set of accepting states. 
\end{definition}

The semantics of finite automata are defined over finite words produced by durational Kripke structures (see Def.~\ref{def:DKS}). In this work, the alphabet $\Sigma$ is chosen to be $2^\AP \times 2^\AP$.
A tuple $\tau=(q_1,(\sigma_1,\sigma_2),q_2) \in \delta$ corresponds to a transition labeled with $(\sigma_1,\sigma_2) \in 2^\AP \times 2^\AP$ from $q_1$ to $q_2$.
A run $\rho$ of a~timed automaton over a finite word $w= \ell_0,\ldots, \ell_n$ is a~sequence $q_0, \ldots,q_n$ of states, such that $q_0 = q_{init}$, and there exists a transition $(q_i,(\ell_i,\ell_{i+1}),q_{i+1}) \in \delta$, for all $0\leq i \leq n-1$. 
A word $w$ is accepted iff there exists a run $\rho = q_0, \ldots, q_n$ over $w$, such that $q_n \in F$ and rejected otherwise. $L(\A)$, called as the language of $\A$, is the set of all words accepted by $\A$. 

An automaton $\A$ is called \emph{non-blocking} if, for all $q\in Q$, and $\ell_1,\ell_2 \in \Sigma$, there exists a transition $(q,(\ell_1,\ell_2),q') \in \delta$. Let us note that every blocking automaton can be trivially converted to a non-blocking automaton by adding transitions to a new state $q_{new} \notin F$.

\subsection{Finite LTL}
Finite automata can capture a large class of properties that are exhibited by traces of a transition system.
However, {some specification languages} with similar expressive power, such as regular expressions or {variants of Linear Temporal Logic (LTL) interpreted over finite runs}, provide a more user-friendly means to express these properties \jana{(see~\cite{sipser, principles} for details).}
We demonstrate in Sec.~\ref{sec:experiments}, how rules of the road and safety rules can be conveniently captured by a slight modification of Finite LTL~\cite{fltl} without the next operator, \jana{called FLTL$_{- \Next}$ and defined below.}

{\begin{definition}[FLTL$_{-\Next}$]
A FLTL$_{-\Next}$ formula $\phi$ over the set of atomic propositions $\AP$ is defined
  inductively as follows:
   \begin{enumerate}
  \setlength{\itemsep}{1pt}
  \setlength{\parskip}{0pt}
  \setlength{\parsep}{0pt}
  \item every pair of atomic propositions, $(\prop,\prop') \in \AP \times \AP$ is a formula,
  \item if $\phi_1$ and $\phi_2$ are formulas, then $\phi_1 \vee    \phi_2$, $\neg \phi_1$, $\phi_1\,\Until\,\phi_2$, $\Always \, \phi_1$, and $\Event \, \phi_1$
    are each formulas,
  \end{enumerate}
 where $\neg$ (negation) and $\vee$
  (disjunction) are standard Boolean connectives, and $\Until$, $\Always$, and $\Event$ are temporal operators.
  \end{definition}}

Unlike the well-known \jana{standard} LTL (see e.g., \cite{principles}), FLTL$_{-\Next}$ is interpreted over finite traces, as those generated by the durational Kripke structure from Def.~\ref{def:DKS}. Informally, $(\prop,\prop')$ holds true on a trace $\ell_0,\ell_1,\ldots,\ell_n$ if $\prop \in \ell_0$, and $\prop' \in \ell_1$. The formula $\phi_1 \,\Until\,\phi_2$ states that there is a future moment when formula $\phi_2$ is true, and formula $\phi_1$
is true at least until $\phi_2$ is true. The formula
$\Always\,\phi$ states that formula $\phi$ holds at all positions
of a finite trace, and $\Event\,\phi$ states that $\phi$ holds at some
future time instance.
%
An FLTL$_{-\Next}$ formula can also be algorithmically translated into a finite automata~\cite{fltl2}. 

\subsection{Level of Unsafety}
Let $\A$ be the automaton for a safety rule with priority $\priority(\A)$. The priority function $\varpi : \A \to \Nat$ assigns priorities to each rule $\A$.
\pcmargin{We assume here that an empty trace by convention always satisfies the safety rule given by any $\A$.}{}


\begin{definition}[Level of Unsafety for a safety rule]
\label{defn:level}
\pcmargin{Let $w = \ell_0, \dots, \ell_n$ be a word over $2^\AP$, for any index set $I = \{i_1,\ldots, i_k\} \subset \{ 0, \ldots n \}$, define}{}
\begin{align*}
\jana{\mathrm{vanish} (w,\{i_1,\ldots i_k\})  = \ell_0, \dots, \ell_{i_j-1},\ell_{i_j+1}, \ldots \ell_n,}
\label{def:level}
\end{align*}
where $1\leq j \leq k$, \ie the finite sequence obtained from $w$ by erasing states indexed with $i_1,\ldots, i_k$.
The level of unsafety $\lambda(w,\A)$ of $w$ with respect to a safety rule expressed as a finite automaton $\A$ is,
\begin{align*}
\lambda(w,\A) =  \min_{I \mid \, \mathrm{vanish}(w,I)\in L(\A)} \ \sum_{i \in I} \priority(\A).
\end{align*}
The level of unsafety for a timed word  $\omega(x) = (\ell_0, d_0), (\ell_1, d_1),\ldots,(\ell_{n-1}, d_{n-1}),(\ell_n,d_n)$ produced by a trajectory $x$ of the dynamical system is,
$$
\hspace{-0.07in} \lambda(x,\A) =  \min_{I \mid \mathrm{vanish}(w(x),I) \in L(\A)} d_i \cdot \priority(\A).
$$
For a trace $r = s_0,\ldots, s_{n+1}$ of the Kripke structure $\K$, it is
$$
\hspace{-0.05in} \lambda(r,\A) =  \min_{I \mid \mathrm{vanish}(w(r),I) \in L(\A)} \sum_{i \in I} \dt(s_i,s_{i+1}) \priority(\A).
$$
\end{definition}
\pcmargin{}{lots of notation changes in this definition}
Consider a sequence of non-empty sets of safety rules $\spec = (\Psi_1 ,\ldots, \Psi_n)$ with each rule $\psi_j \in \Psi_i$, for all $1 \leq i \leq n$ given in the form of a finite automaton $\A_{i,j}$.
The ordered set $\spec$ together with the priority function~$\priority$ is called a set of safety rules with priorities $(\spec, \priority)$.
We now extend the definition of the level of unsafety for a  word $w$ and a trace $r$ to a set of safety rules with priorities $(\spec,\priority)$ as follows.

\vspace{-0.05in}
\begin{definition}[Level of Unsafety for a set of rules]
\label{defn:level_unsafety_set_of_rules}
The level of unsafety of a word with respect to a set of rules $\Psi_i$, $\lambda(w,\Psi_i)$ and the level of unsafety with respect to a set of rules with priorities $(\spec,\priority)$ are defined as,
\begin{align*}
&\lambda(w,\Psi_i) = \sum_{\A_{i,j} \in \Psi_i} \lambda(w,\A_{i,j}), \\
& \lambda(w,\spec) = \big(\lambda(w,\Psi_1),\ldots, \lambda(w,\Psi_n)\big)
\end{align*}
Level of unsafety for \jana{a trajectory of the dynamical system and} a trace $r$ of $\K$ with respect to a set of rules with priorities is defined similarly. \pcmargin{The standard lexicographic ordering is used to compare the level of unsafety of two traces $r_1$, $r_2$.}{}
\end{definition}

\myclearpage
\section{Problem Formulation}
\label{sec:problem_formulation}

For a compact set $\mathcal{S} \subset \Real^d$, define $s_{init} \in \mathcal{S}$ to be the initial state and a compact subset $\mathcal{S}_{goal} \subset \mathcal{S}$ as the goal region.
Given the dynamical system in Eq.~\eqref{eqn:dynamics}, define a task specification $\task$ to be, ``traveling from $s_{init}$ to  $\mathcal{S}_{goal}$''.
%
\pcmargin{The word produced by a trajectory $x : [0, T] \to X$, $w(x) = \ell_0, \ell_1, \ldots, \ell_n$ is said to satisfy  the task $\task$ if $\ell_0 = s_{init}$ and $\ell_n \in \mathcal{S}_{goal}$. Similarly, a trace of the Kripke structure, $r = s_0, \ldots, s_n$ satisfies $\task$ if $s_0 = s_{init}$ and $s_n \in \mathcal{S}_{goal}$. We assume in this work that this task is feasible.}{removed the old notion of a continuous trajectory satisfying $\task$}

\vspace{-0.02in}
\begin{problem}
\label{prob:dynamical_form}
Given a dynamical system as shown in Eq.~\eqref{eqn:dynamics}, a task specification $\task$, a set of safety rules with priorities $(\spec,\priority)$ and a continuous function $\cost(x)$ that maps a trajectory $x$ of the dynamical system to a non-negative cost, find a trajectory $x^* :[0,T] \to X$ producing a timed word $\w(x^*)=(\ell_0,d_0)\ldots (\ell_n,d_n)$ and a word $w(x)$ such that,
\begin{enumerate}[(i)]
\item $w(x)$ satisfies the task specification $\task$,
\item $x^*$ minimizes the level of unsafety, \jana{$\lambda(x',\spec)$}, among all trajectories $x'$ that satisfy condition (i),
\item $x^*$ minimizes $\cost(x'')$ among all trajectories $x''$ that satisfy conditions (i) and (ii).
\end{enumerate}
\end{problem}
The solution of this problem as defined above exists if the task $\task$ is feasible. In this work, we restrict ourselves to minimum-time cost functions, i.e., $\cost(x) = \int_0^T 1dt$. The algorithm described here however applies to a much wider class of functions including discounted cost as well as state and control based cost functions with minor changes. In order to develop an algorithmic approach for Prob.~\ref{prob:dynamical_form}, we convert it to the following problem defined on a trace-inclusive durational Kripke structure. Thm.~\ref{thm:convergence} connects the solutions of Prob.~\ref{prob:dks_form} to those of Prob.~\ref{prob:dynamical_form}.

\vspace{-0.05in}
\begin{problem}
\label{prob:dks_form}
Given a durational Kripke structure $\kripke$ that is trace-inclusive for the dynamical system in Eq.~\eqref{eqn:dynamics}, a task specification $\task$, a set of safety rules with priorities $(\spec, \priority)$ and a cost function $\cost(x)$, find a 
finite trace $r^* = s_0, s_1, \ldots, s_n$ of $\K$ such that,
\begin{enumerate}[(i)]
\item $r^*$ satisfies $\task$,
\item $r^*$ minimizes $\lambda(r',\spec)$ among all traces $r'$  of $\K$ that satisfy condition (i),
\item $r^*$ minimizes $\ag{r}$ among all traces $r''$ satisfying (i), (ii).
\end{enumerate}
\end{problem}
\pcmargin{}{removed references to the continuous trajectory $x^*$ from here to make the presentation cleaner}

\myclearpage
\vspace{-0.05in}
\section{Algorithm}
\label{sec:algorithm}
This section describes an algorithm for finding minimum-constraint violation trajectories for a dynamical system. 
We then propose an algorithm, based on RRT$^*$, to incrementally construct a product of the Kripke structure and automata representing safety rules. Roughly, the shortest path in the product uniquely maps to a trace of the Kripke structure that minimizes the level of unsafety.  Let us note that the algorithm returns a trajectory that satisfies all rules and minimizes the cost function if it is possible to do so.

\subsection{Weighted Product Automaton}
First, we augment each automaton $\A_{i,j} \in \spec$ with new transitions and weights, such that the resulting weighted automaton also accepts all words $w$ that {\em do not} satisfy the rule $\A_{i,j}$; the weights are picked such that the weight of an accepting run over $w$ determines the level of unsafety of $w$ with respect to $\A_{i,j}$  (see Def.~\ref{defn:automaton_one}).
Second, we combine all the weighted automata into a single weighted automaton $\allA$; the weights of this automaton capture the level of unsafety with respect to a set of safety rules with priorities $(\spec,\priority)$ (see Def.~\ref{defn:automaton_two}). 
Third, we build the product of the durational Kripke structure $\K$ and the automaton $\allA$ (see Def.~\ref{defn:automaton_product}); weights of this product correspond to the level of unsafety of traces of $\K$.

We now proceed to describe each of these steps in detail and summarize the purpose of each construction in a lemma (see Def.~\ref{def:modified_A_small}--\ref{def:prod_big_weighted} and Lem.~
\ref{lemma:1}--\ref{lemma:3}). The material presented in this section is a slight modification of our earlier algorithm for finding a trace of a weighted transition system that minimizes the level of unsafety~
{\cite{hscc2013}}. For the sake of brevity, proofs of these lemmas are omitted and can be found in~{\cite{hscc2013}}.

\begin{definition}[Weighted Automaton] 
\label{defn:automaton_one}
\label{def:modified_A_small}
For a non-blocking finite automaton $\autom$, the weighted finite automaton is defined as
$
\overline\A = (Q, q_{init}, 2^\AP, \overline\delta, F, \overline \W),
$ 
where, $\overline{\delta} = \delta \cup \{(q, (\sigma,\sigma'),q') \mid q, q' \in Q$, $(\sigma, \sigma') \in {2^\AP}^2 \}$,
$$
  \overline\W\big(\tau)= \left\{ 
  \begin{array}{l l}
    0 & \text{if $\tau \in \delta$} \\
    \priority(\A) & \text{if $\tau \in \overline{\delta} \setminus \delta.$}
  \end{array} \right.
$$
\end{definition}
\begin{lemma}
For a rule $\psi_{i,j}$ given as an automaton $\A_{i,j}$, any word over $2^\AP$ is accepted by $\overline A_{i,j}$ and the weight of the shortest accepting run is equal to $\lambda(w, \psi_{i,j})$.
\label{lemma:1}
\end{lemma}
A single weighted automaton $\overline \A_\spec$ is created by combining all automata $\overline \A_{{i,j}}$, where $\A_{i,j} \in \Psi_i \in \spec$. This captures the level of unsafety with respect to the whole set of safety rules with priorities $(\spec,\priority)$ through its weight function.

\begin{definition}[Automaton $\allA$] 
\label{defn:automaton_two}
\label{def:modified_A_big}
The  weighted automaton $\overline\A_\spec = (\overline Q,\overline q_{init},2^\AP,C,\overline \delta,\overline F, \overline \W)$ is defined as follows:
\begin{itemize}\itemsep0ex
\item $\overline{Q} = Q_{1,1}\ldots \times \ldots Q_{1,m_1} \ldots \times \ldots Q_{n,1} \ldots \times \ldots Q_{n,m_n}$;
\item $\overline{q}_{init} = (\overline q_{init,1,1},\ldots,\overline q_{init,n,m_n})$;
\item $(p,(\sigma,\sigma'),p') \in \overline{\delta}$ if~
\begin{itemize}
\item $p=(q_{1,1},\ldots,q_{n,m_n})$, $p'= (q_{1,1}',\ldots,q_{n,m_n}')$, and 
\item $(q_{i,j},(\sigma,\sigma'),q_{i,j}') \in \overline\delta_{i,j}$, for all $i \in \{1,\ldots,n\}, j\in \{1,\ldots m_i\}$.
\end{itemize}
Also, $\overline \W((p,(\sigma,\sigma'),p')) = (x_1,\ldots,x_n)$, where $x_i = \sum_{j=1}^{m_i} \overline \W_{i,j}(q_{i,j},(\sigma,\sigma'),q_{i,j}')$;
\item $\overline F = \{(q_{1,1},\ldots,q_{n,m_n}) \mid q_{i,j} \in \overline F_{i,j}, \text{ for all } \\ i\in \{1,\ldots,n\}, j\in \{1,\ldots m_i\}\}$
\end{itemize}
\end{definition}

\begin{lemma}
Any word $w$ over $2^\AP$ is accepted by $\allA$ and the weight of the shortest accepting run of $\allA$ over $w$ is equal to the level of unsafety $\lambda(w,\spec)$.
\label{lemma:2}
\end{lemma}

\begin{definition}[Weighted Product Automaton $\P$] 
\label{defn:automaton_product}
\label{def:prod_big_weighted}
We build the weighted product automaton,
$$\P = \K \ \otimes \ \allA = (Q_\P,q_{init,\P},\delta_\P,F_\P,\W_\P)$$ 
of the Kripke structure $\kripke$ and the augmented automaton $\overline\A_\spec = (\overline Q,\overline q_{init},2^\AP,\overline \delta,\overline F, \overline \W)$ as, 
\begin{itemize}\itemsep0ex
\item $Q_\P = S \times \overline Q $ is a set of states; 
\item $q_{init,\P} = (s_{init},\overline q_{init})$ is the initial state; 
\item $\delta_\P \subseteq Q_\P \times Q_\P$ is a non-deterministic transition relation, where $((s,q),(s',q')) \in \delta_\P$ if $(s,s')\in \R$, and
there exists a transition $(q,(\L(s),\L(s')), q') \in \overline \delta$.
Then also,
$$\W_\P \big((s,q),(s',q')\big) = (x_1 \cdot \dt(s,s'), \ldots, x_n \cdot \dt(s,s') ),$$
where $(x_1,\ldots,x_n)=\overline \W(q,\L(s),\L(s'),q') $ and,
\item $F_\P =  (S \cap \mathcal{S}_{goal}) \times \overline F$ is a set of accepting states.
\end{itemize}
\end{definition}

%
A product automaton is in fact, a finite automaton extended with weights. A run of a product automaton is a sequence $\rho = p_0,\ldots,p_n$, such that $p_0 = q_{init,\P}$, and $(p_i,p_i+1) \in \delta_\P$, for all $0\leq i <n$ and it is accepting if $p_n \in F_\P$. The weight of a run $\W_\P(\rho)$ is the tuple obtained by component-wise sum of the weights associated with the transitions executed along the run. The \emph{shortest run over $w$} is then a run $\rho$ minimizing the weight $\W_\P(\rho)$ in the lexicographical ordering.

\begin{lemma}
The shortest accepting run (in the lexicographical ordering with respect to $\W_\P$), $p_0\ldots p_n$ of $\P$ from the state $p_0 = q_{init,\P}$ to a state $p_n \in F_\P$ projects onto a trace $r = s_0, \ldots s_n$ of $\K$ that minimizes the level of unsafety.
\label{lemma:3}
\end{lemma}

\subsection{Incremental Weighted Product Automaton}

In this section, we incrementally construct the weighted product automaton (see Def.~\ref{defn:automaton_product}) and maintain the trace that minimizes the level of unsafety for a set of safety rules $\spec$. A few preliminary procedures of the algorithm are as follows :

\subsubsection{Sampling}
The ${\tt Sample}$ procedure samples an independent, identically distributed state $s$ from a uniform distribution supported over the bounded set $\mathcal{S}$.
\subsubsection{Nearest neighbors}
The ${\tt Near} $ procedure returns the set,
$$ S_{near}(s) = \{ s' | \norm{s' - s}_2 \leq \gamma \left(\log n/n \right)^{1/d};\ s' \in S \}$$ 
where $n = |S|$ and $\gamma$ is a constant given in Thm.~\ref{thm:convergence}. 
\subsubsection{Steering}
Given two states $s, s'$, the ${\tt Steer}(s', s)$ procedure computes the pair $(x, T)$ where $x : [0, T] \to X$ is a trajectory such that, (i) $x(0) = s'$, (ii) $x(T) = s$ and, (iii) $x$ minimizes the cost function $\cost(x) = T$. 
If a trajectory $x$ is found, return true, else return false.
\subsubsection{Connecting}
For a state $s' \in S_{near}$, if ${\tt Steer}(s', s)$ returns true, for all nodes $z' = (s', q') \in Q_\P$, for all $(z', (s, q)) \in \delta_\P$, the procedure ${\tt Connect}(s', s)$ adds the state $z = (s, q)$ to the set $Q_\P$, adds $(z', z)$ to $\delta_\P$ and calculates $\W_\P(z', z)$. If $s \in \mathcal{S}_{goal}$ and $q \in F$, it adds $(s, q)$ to $F_\P$.
\subsubsection{Updating costs}
The procedure ${\tt Update}(s)$ updates the level of unsafety $\Ja(z)$ and the cost $\Jt(s)$ from the root for a node $z = (s, q)$ as shown in Alg.~\ref{alg:update} using the sets,
$$S_{steer}(s) = \{ s'\ |\ s' \in S_{near}(s);\ {\tt Steer}(s', s) \text{ returns }{\bf true} \},$$
$$Z_{steer}(s) = \{ (s', q')\ |\ s' \in S_{steer}(s);\ (s',q') \in Q_\P \}.$$
\subsubsection{Rewiring}
In order to ensure asymptotic optimality, the $\tt Rewire$ procedure recalculates the best parent ${\tt Par}(s')$ for all states $s' \in S_{near}(s)$ as shown in Alg.~\ref{alg:rewire}. The complexity of this procedure can be reduced by noting that $s'$ only needs to check if the new sample can be its parent by comparing costs $\Ja, \Jt$, otherwise its parent remains the same.

Finally, Alg.~\ref{alg:inc_product} creates the weighted product automaton as defined in Def.~\ref{defn:automaton_product} incrementally. It also maintains the best state $z^* = (s^*, q^*) \in F_\P$. The trace $r^* = s_0, s_1, \ldots, s_n$ of the Kripke structure $\K$ that minimizes the level of unsafety and is a solution to Prob.~\ref{prob:dks_form} can then be obtained from $z^*$ by following ${\tt Par}(s^*)$. Since $\K$ is trace-inclusive, the continuous-time trajectory $x^*$ can be obtained by concatenating smaller trajectories.
Let $(x_i, T_i)$ be the trajectory returned by ${\tt Steer}(s_i, s_{i+1})$ for all states $s_i \in r^*$. The concatenated trajectory $x^* : [0, T] \to X$ is such that $T = \sum_{i=0}^{n-1} T_i$ and $x_n(t + \sum_{k=0}^{i-1} T_k) = x_i(t)$ for all $i < n$.

\IncMargin{0.04in}
\begin{algorithm}[!h]
\small
Input : $n,\ \S,\ \overline{\A}_\spec$\;
$\P \la \varnothing$; $Q_\P \la q_\P^{init}$; $\Ja(s_{init}) \la 0; \Jt(s_{init}) \la 0$\; 
$i \la 0$\;
\For{$i \leq n$}
{
	$s \la {\tt Sample}$\;
	\For{$s' \in {\tt Near}(s)$}
	{
		\If{${\tt Steer}(s', s)$}
		{
			${\tt Connect}(s', s)$\;
		}
	}
	${\tt Par}, \Ja, \Jt \la {\tt Update}(s)$\;
	$\P, \Ja, \Jt \la {\tt Rewire}(s)$\;
}
$\P_n \la (Q_\P, q_\P^{init}, \delta_\P, F_\P, \W_\P)$\;
\Return {$\P_n$}
\caption{$\tt Create\ Product$}
\label{alg:inc_product}
\end{algorithm}
\DecMargin{0.04in}
\IncMargin{0.04in}
\begin{algorithm}[!h]
\small
\For{$z = (s, q) \in Q_\P$}
{
	$\displaystyle \Ja(z) \la \min_{z' \in Z_{steer}}\ \W_\P(z', z) + \Ja(z')$\;
	$\displaystyle Z^* \leftarrow \arg \min_{z' \in Z_{steer}}\ \W_\P(z', z) + \Ja(z')$\;
	$\displaystyle \Jt(s) \leftarrow \min_{z' \in Z^*} \cost(s',s) + \Jt(s')$\;
	$\displaystyle {\tt Par}(z) \leftarrow \arg \min_{z' \in Z^*} \cost(s',s) + \Jt(s')$\;
}
\Return{${\tt Par}, \Ja, \Jt$}
\caption{${\tt Update} (s, \P)$}
\label{alg:update}
\end{algorithm}
\DecMargin{0.04in}

\IncMargin{0.04in}
\begin{algorithm}[!h]
\small
\For{$s' \in S_{steer}(s)$}
{
	\If{ ${\tt Steer}(s, s')$}{
		${\tt Connect}(s, s')$\;
	}
	$\Ja, \Jt \la {\tt Update}(s')$\;
}
\Return{$\P$}
\caption{${\tt Rewire}(s, \P)$}
\label{alg:rewire}
\end{algorithm}
\DecMargin{0.04in}

\myclearpage
\section{Analysis}
\label{sec:analysis}
In this section, we analyze the convergence properties of Alg.~\ref{alg:inc_product}. In particular, we prove that the continuous-time trajectory $x_n$ given by the algorithm after $n$ iterations converges to the solution of Prob.~\ref{prob:dynamical_form} as the number of states in the durational Kripke structure $\K_n$ goes to infinity, with probability one. A brief analysis of the computational complexity of the algorithm is also carried out here. Due to lack of space, we only sketch the proofs.

\begin{theorem}
\label{thm:convergence}
The probability that Alg.~\ref{alg:inc_product} returns a durational Kripke structure $\K_n$ and a trajectory of the dynamical system $x_n$, that converges to the solution of Prob.~\ref{prob:dynamical_form} in the bounded variation norm sense, approaches one as the number of states in $\K_n$ tends to infinity, i.e.,
$$
\PP \left( \{ \lim_{n \to \infty}\ \norm{x_n - x^*}_{BV} = 0 \} \right) = 1
$$
\end{theorem}
\vspace{-0.03in}
\begin{proof}(Sketch) The proof primarily follows from the asymptotic optimality of the RRT$^*$ algorithm (see Theorem 34 in~\cite{karaman2011sampling}). Let $x^* : [0, T] \to X$ be the solution of Prob.~\ref{prob:dynamical_form} that satisfies the task $\task$ and minimizes the level of unsafety. For a large enough $n$, define a finite sequence of overlapping balls $B_n = \{ B_{n,1}, \ldots, B_{n,m} \}$ around the optimal trajectory $x^*$. The radius of these balls is set to be some fraction of $\gamma (\log n/n)^{1/d}$ such that any point in $s \in B_{n, m}$ can connect to any other point $s' \in B_{n, m+1}$ using the ${\tt Steer}(s, s')$ function. It can then be shown that each ball in $B_n$ contains at least one state of $\K_n$ with probability one. In such a case, there also exists a trace $r_n = s_0, s_1, \ldots, s_n$ of $\K_n$ such that every state $s_i$ lies in some ball $B_{n, m}$. Also, for a large enough $n$, the level of unsafety of $r_n$, $\lambda(r_n, \spec)$ is equal to the level of unsafety of the word generated by the trajectory $x^*$, $\lambda(\w(x^*), \spec)$, i.e., \ALG returns the trace with the minimum level of unsafety among all traces of the Kripke structure $\K$ satisfying the task $\phi$. Finally, it can be shown that the trajectory $x_n$ constructing by contanetating smaller trajectories joining consecutive states of $r$, i.e., $s_0, s_1, \ldots$ converges to $x^*$ almost surely as $n \to \infty$.

In this proof, $\gamma > 2 \left(2 + 1/d \right)^{1/d} \left(\mu(\cal{S})/\zeta_d \right)^{1/d}$, where $\mu(\cal{S})$ is the Lebesgue measure of the set $\cal{S}$ and $\zeta_d$ is the volume of the unit ball of dimensionality $d$.
\end{proof}

The following lemma is an immediate consequence of Thm.~\ref{thm:convergence} and the continuity of the cost function $c(x)$.
\begin{lemma}
\label{lem:cost_convergence}
The cost of the solution converges to the optimal cost, $c^* = c(x^*)$, as the number of samples approaches infinity, almost surely, i.e,
$
\PP \left( \{ \lim_{n \to \infty}\ c(x_n) = c^* \}\right) = 1.
$
\end{lemma}

%
Let us now comment on the computational complexity of MVRRT$^*$.
%
%
%
Note that there are an expected $\cal{O}(\log n)$ samples in a ball of radius $\gamma (\log n/n)^{1/d}$. The procedure ${\tt Steer}$ is called on an expected $\cal{O}(\log n)$ samples while because the automaton $\allA$ is non-deterministic, the procedure ${\tt Connect}$ adds at most $m^2$ new states in the product automaton per sample. The procedure $\tt Update$ requires at most $\cal{O}(m^2 \log n)$ time call. The $\tt Rewire$ procedure simply updates the parents of the $\cal{O}(\log n)$ neighboring samples which take $\cal{O}(m^2 \log n)$ time. In total, the computational complexity of \ALG is $\cal{O}(m^2 \log n)$ per iteration.


\myclearpage

\newcommand{\rl}{{rl}}
\renewcommand{\ll}{{ll}}
\newcommand{\side}{{sw}}
\newcommand{\obs}{{obs}}
\newcommand{\dir}{{dir}}
\newcommand{\dotted}{{dotted}}
\newcommand{\solid}{{solid}}

\section{Experiments} 
\label{sec:experiments}

In this section, we consider an autonomous vehicle modeled as a Dubins car in an urban environment with road-safety rules and evaluate the performance of \ALG in a number of different situations.

\subsection{Experimental Setup}
Consider a Dubins car, \ie a curvature-constrained vehicle with dynamics, $\dot{x} = v \cos(\theta), \dot{y} = v \sin(\theta)$ and $\dot{\theta} = u$.
The state of the system is the vector $[ \, x, \, y, \, \theta]^T$, and the input is $u(t)$, where $|u(t)| \leq 1$ for all $t \geq 0$. The vehicle is assumed to travel at a constant speed $v$. As shown in~\cite{dubins1957curves}, time-optimal trajectories for this system in an obstacle-free environment can be easily calculated.

\begin{figure}
\centering
\includegraphics[width=0.3 \textwidth]{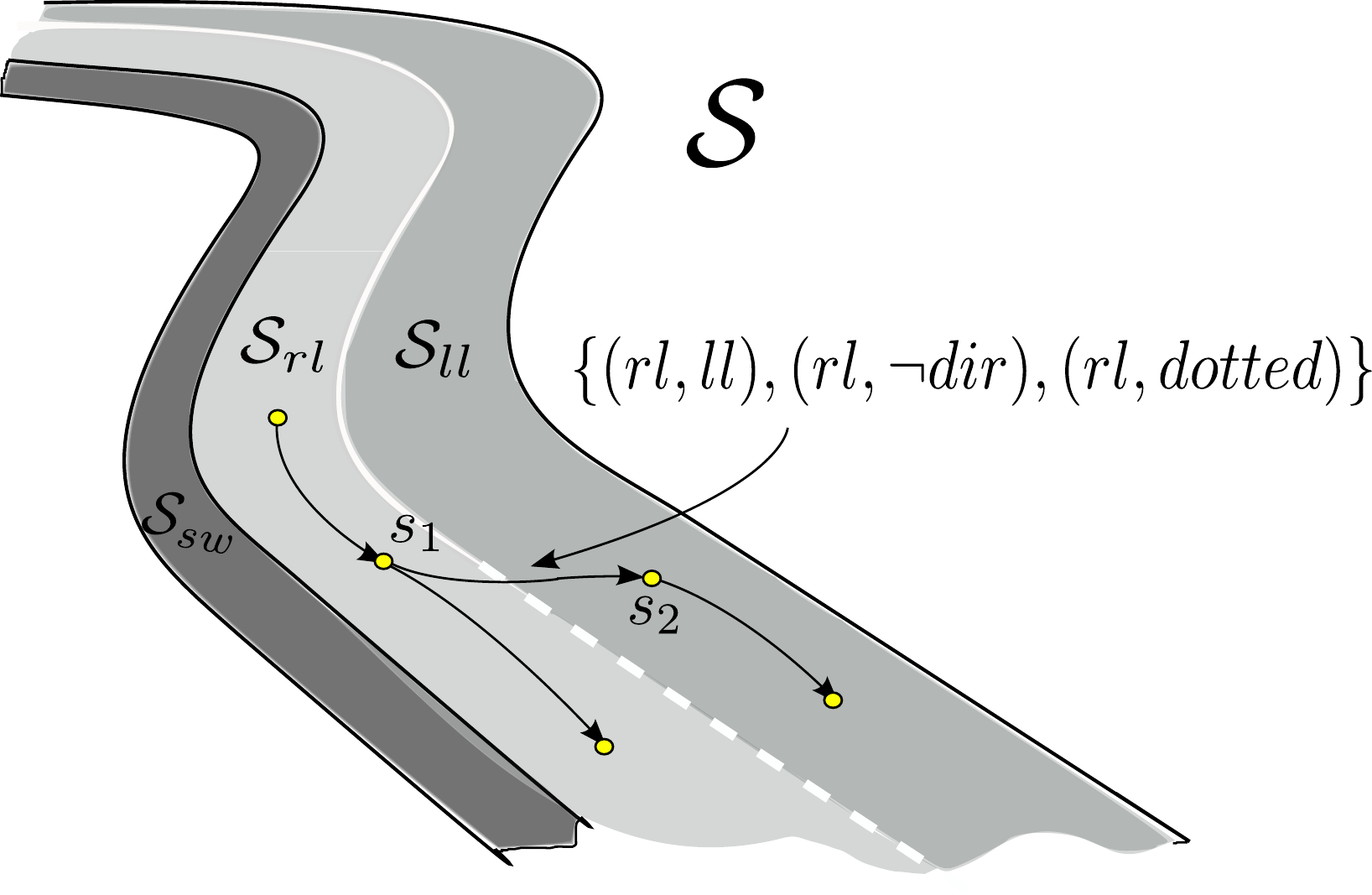}
\caption{Partitions of the working domain $\S$. The transition from $s_1$ to $s_2$ is labeled with, for example, $\{ (\rl, \ll), (\rl, \neg \dir), (\rl, \dotted)\}$.}
\label{fig:partitions}
\end{figure}

%
We partition the working domain $\S$ into compact non-empty subsets $\S_{\obs}$ which is the union of obstacled regions, $\S_{\side}$ which represents the sidewalk and $\S_{\rl}$, $\S_{\ll}$ which are the right and left lanes, respectively, as illustrated in Fig.~\ref{fig:partitions}. $\S_\obs$ is empty if there are no obstacles.
Based on this partitioning, we define the set of atomic propositions as,
$ 
\Pi = \{ \side, \rl, \ll, \dir, \dotted, \solid \}.
$
A proposition $p \in \{ \side, \rl, \ll \}$ is true at a state $s \in S$, if $s \in \S_p$ with $\rl, \ll$ being mutually exclusive. $\dir$ is true iff the heading of the car is in the correct direction, i.e., if $s$ is such that the car heading forwards and $\rl$ is true. Atomic propositions, $\dotted$ and $\solid$, depict the nature of lane markers. Note that obstacles are not considered while constructing $\Pi$ since we do not 
desire a trajectory that goes over an obstacle. The $\tt Steer$ procedure in Sec.~\ref{sec:algorithm},  instead, returns false if any state along the trajectory lies in $\S_{obs}$. This change does not 
affect the correctness and the overall complexity of MVRRT$^*$. 

\subsection{Safety Rules}
Given a task $\task$ such as finding a trajectory from $s_{init}$ to the goal region $\S_{goal}$, we require the vehicle to follow the following rules: \textit{(i)} do not travel on sidewalks (\emph{sidewalk rule}), \textit{(ii)} do not cross solid center lines (\emph{hard lane changing}), \textit{(iii.a)} always travel in the correct direction (\emph{direction rule}), \textit{(iii.b)} do not cross dotted center lines (\emph{soft lane changing}).

We describe the rules with the following FLTL$_{-\Next}$ formulas and corresponding finite automata in Fig.~\ref{fig:rule1}. Note that we use 2-tuples of atomic propositions from $\Pi$ as the alphabet for both formulas and the automata, to specify not only properties of individual states, but also of transitions. The two components capture the atomic propositions of the starting and the ending state respectively.
\subsubsection*{(i) Sidewalk} Do not take a transition that ends in $\mathcal{S}_{sw}$.
$$
\psi_{1,1} = \Always\ \bigwedge_{* \in 2^\AP} \neg (*, \side)
$$
\subsubsection*{(ii) Hard lane change} Do not cross a solid center line.
$$
\psi_{2,1} = \Always \Big(\neg \big((\rl,\solid) \wedge (\rl,\ll)\big) \vee \big((\ll, \solid) \wedge (\ll,\rl)\big)\Big)$$
\subsubsection*{(iii.a) Direction} Do not travel in the wrong direction. 
$$
	\psi_{3,1} = \Always\ \bigvee_{*\in 2^\AP}  (*,\ \dir)
$$
\subsubsection*{(iii.b) Soft lane change} 
\begin{figure}[h!]
\begin{center}
\usetikzlibrary{arrows,positioning,automata,shadows,fit,shapes}
\begin{tikzpicture}[->,>=stealth',shorten >=1pt,auto, node distance=2.5cm,semithick,initial text=]
  \tikzstyle{every state}=[draw]
	  \node[initial, accepting, state, minimum size=4ex] (q1) {$q_1$};	
\begin{scope}[every node/.style={scale=.8}]
  \path (q1) edge [loop right] node {$(\ell,\ell')$} (q1);
 \end{scope}
\end{tikzpicture}
\end{center}
\caption{Rule \textit{iii.b} : For the sake of brevity, the transition above represents all transitions, where 
\textit{(i)}  $\ell, \ell' \subseteq 2^\AP$, such that $\rl \in \ell$ and $\dotted,\ll \in \ell'$, or $\ll \in \ell$ and $\dotted,\rl \in \ell$, and
\textit{(ii)} $\ell, \ell' \subseteq 2^\AP$, such that $\rl \in \ell$ and $\solid,\ll \in \ell'$, or $\ll \in \ell$ and $\solid,\rl \in \ell$.
}
\label{fig:rule1}
\end{figure}
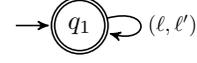

Do not cross a dotted center line.
$$
\psi_{3,2} = \Always \Big(\neg \big((\rl,\dotted) \wedge (\rl,\ll)\big) \vee \big((\ll, \dotted) \wedge (\ll,\rl)\big)\Big)$$
The finite automata for rules \textit{(i)}-\textit{(iii.b)} are all of the same form (see Fig.~\ref{fig:rule1}). 

While it is quite natural to disobey the direction and the soft lane change rules, a solid line should not be crossed.
This gives three different priority classes
$$(\Psi_1, \Psi_2,\Psi_3), \priority) = ((\{\psi_{1,1} \}, \{\psi_{2,1}\},\{\psi_{3,1}, \psi_{3,2}\}) , \priority),$$ where $\priority(\psi_{1,1}) = \priority(\psi_{2,1}) = \priority(\psi_{3,1}) = 1$ and $\priority(\psi_{3,2}) = 10$.
Note that costs for $\psi_{2,1}$ and $\psi_{3,2}$ are incurred only once per crossing and do not depend upon the duration of the transition. Within the third class, we put higher priority on the soft lane change rule to avoid frequent lane switching, for instance in case two obstacles are very close to each other and it is not advantageous to come back to the right lane for a short period of time, e.g., see Fig.~\ref{fig:big_sw_dir}.

\subsection{Simulation Experiments}
\begin{figure}[!htbp]
\centering
\vspace{0.1in}
\subfloat[] {\includegraphics[width= 0.7 \columnwidth]{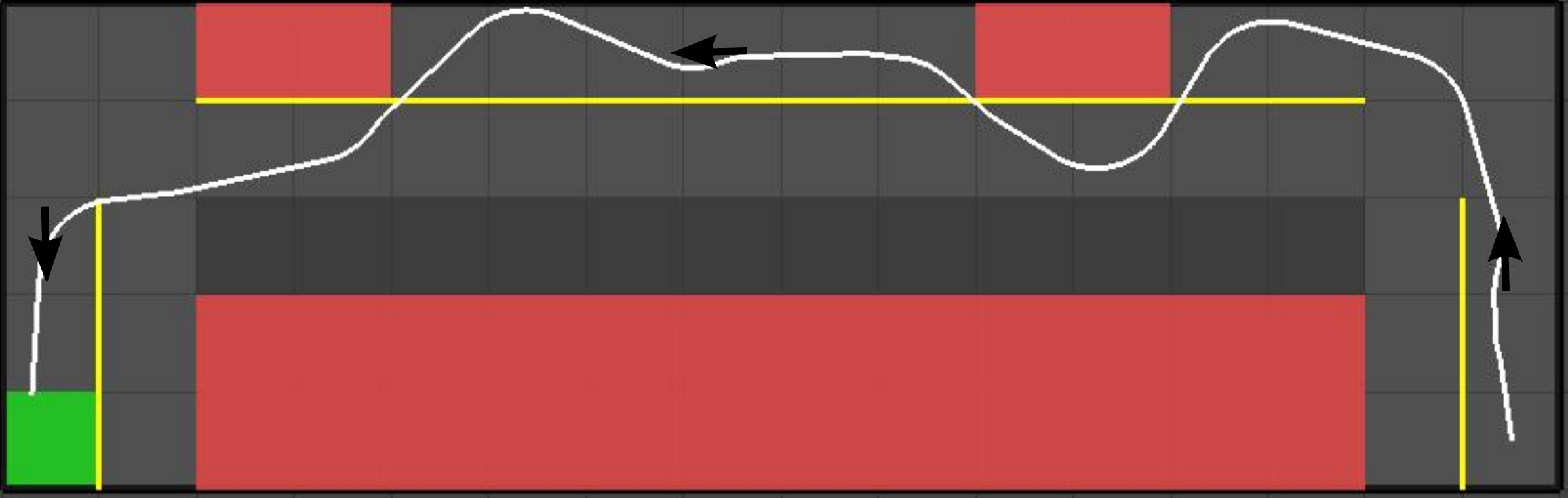} \label{fig:sw_dir}} \\
\subfloat[] {\includegraphics[width= 0.7 \columnwidth]{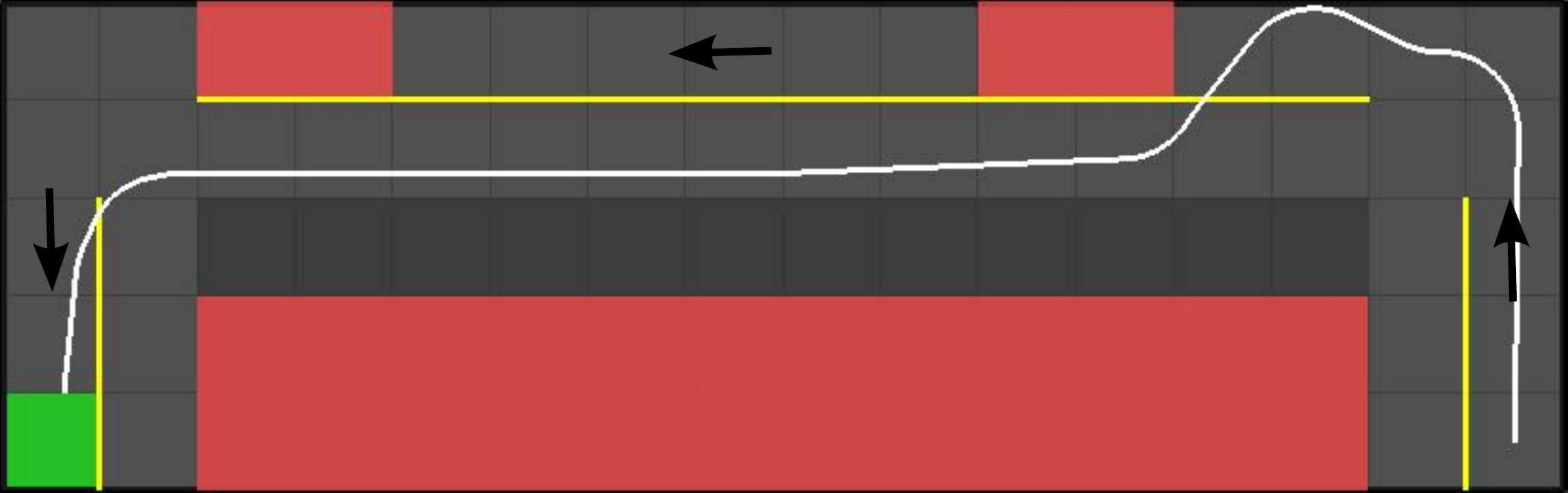} \label{fig:sw_dir_dotted}}
\caption{Fig.~\ref{fig:sw_dir} shows the solution after 60 secs. for sidewalk and direction rules. Upon introducing the soft lane changing rule in Fig.~\ref{fig:sw_dir_dotted}, the vehicle does not return to the right lane after passing the first obstacle. 
}
\label{fig:big_sw_dir}
\end{figure}

\ALG was implemented in C++ on a 2.2GHz processor with 4GB of RAM for the experiments in this section. We present a number of different scenarios in the same environment to be able to quantitatively compare the performance.
In Fig.~\ref{fig:big_sw_dir}, the Dubins car starts from the lower right hand corner while the goal region marked in green is located in the lower left hand corner. Light grey denotes the right and left lanes, $\S_\rl$ and $\S_\ll$. A sidewalk $\S_\side$ is depicted in dark grey. The dotted center line is denoted as a thin yellow line while solid center lines are marked using double lines.
Stationary obstacles in this environment are shown in red.

\begin{figure}[!tbp]
\centering
\includegraphics[width=0.35 \textwidth]{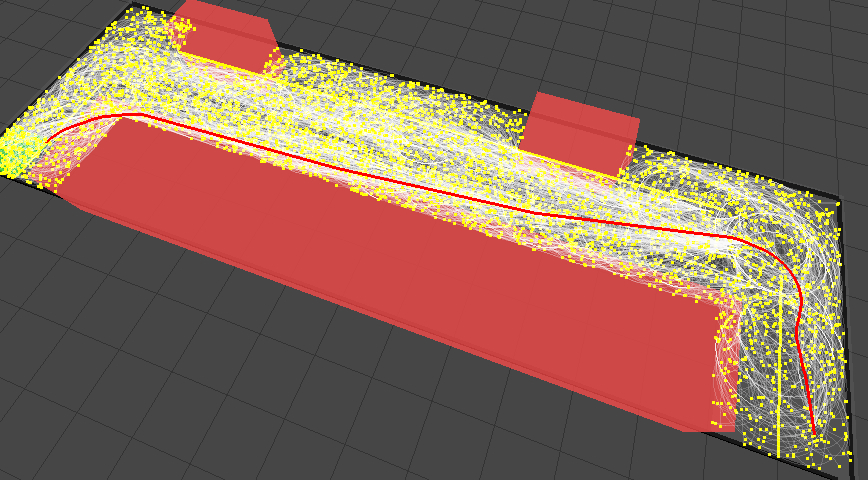}
\caption{\ALG tree after 40 sec. on an example without any safety rules. States of the Kripke structure are shown in yellow while edges are shown in white. The shortest trajectory shown in red to the goal region avoids obstacles but uses the sidewalk.}
\label{fig:rrts}
\end{figure}

\paragraph{Case 1} First, we consider a scenario without any safety or road rules. The \ALG algorithm then simply aims to find the shortest obstacle-free trajectory from the initial state to the goal region. Note, that in this case, \ALG  performs the same steps as the RRT$^*$ algorithm. The solution computed after 40 seconds has a cost of $88.3$ and is illustrated in Fig.~\ref{fig:rrts} together with the adjoining tree.

\paragraph{Case 2} Next, we introduce the sidewalk rule $\psi_{1,1}$ and the direction rule $\psi_{3,1}$. Without any penalty on frequent lane changing, the car goes back into the right lane after passing the first obstacle. It has to cross the center line again in order to pass the second obstacle and reach the goal region. Fig~\ref{fig:sw_dir} depicts the solution that has a cost of $122.3$ along with a level of unsafety of $46.4$ for breaking $\psi_{3,1}$.

Upon introducing the rule $\psi_{3,2}$, the vehicle does not go back into the right lane after passing the first obstacle. Figure~\ref{fig:sw_dir_dotted} shows this solution with a level of unsafety of $84.1$ for breaking both $\psi_{3,1}$ and $\psi_{3,2}$ whereas the level of unsafety in this case for the trajectory in Fig.~\ref{fig:sw_dir} is $87.4$.


\begin{figure}[!h]
\centering
\subfloat[] {\includegraphics[width=0.22 \textwidth]{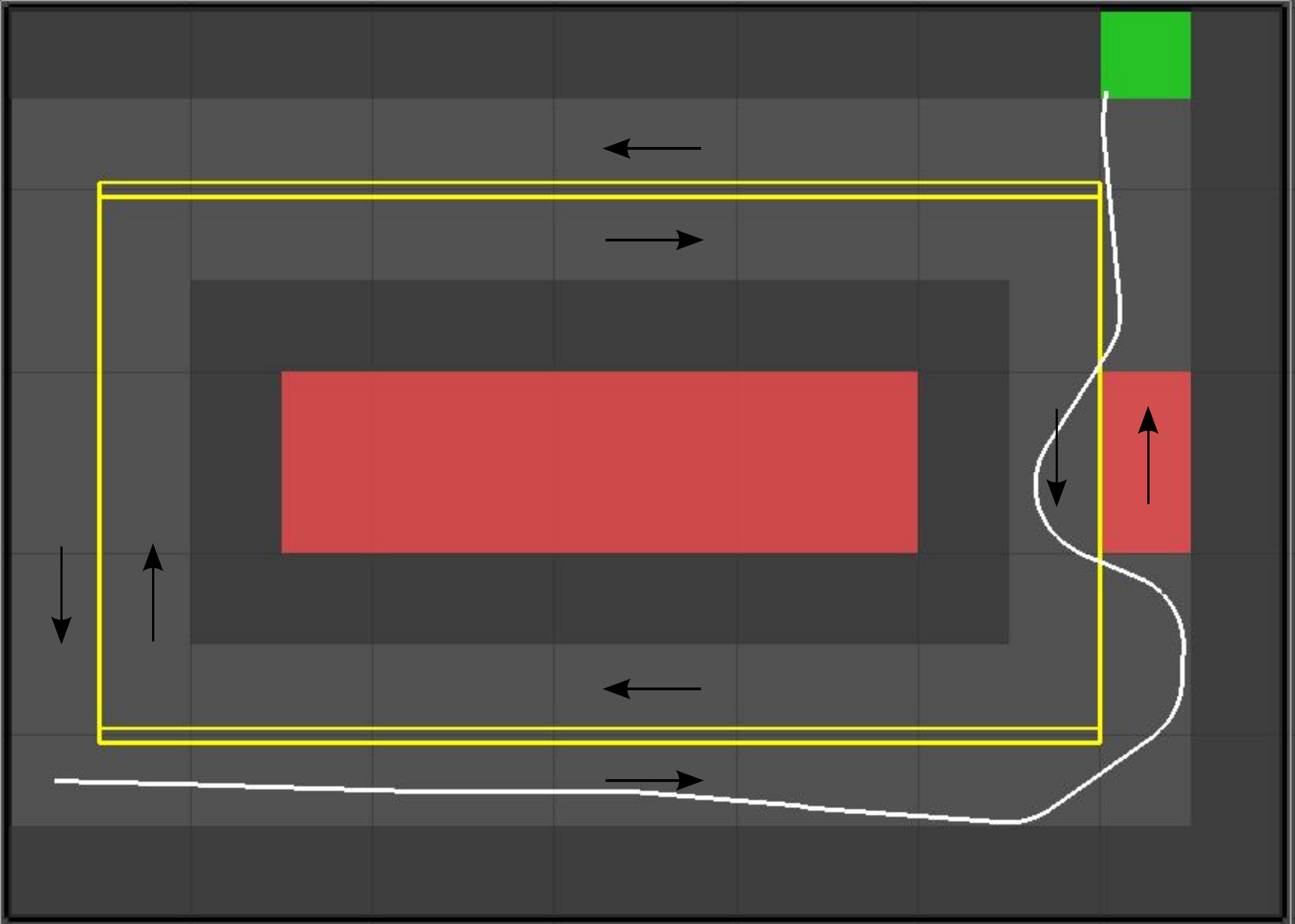}\label{fig:eg2_sw_dir_no_converge}}\hspace{0.05in}
\subfloat[] {\includegraphics[width=0.22 \textwidth]{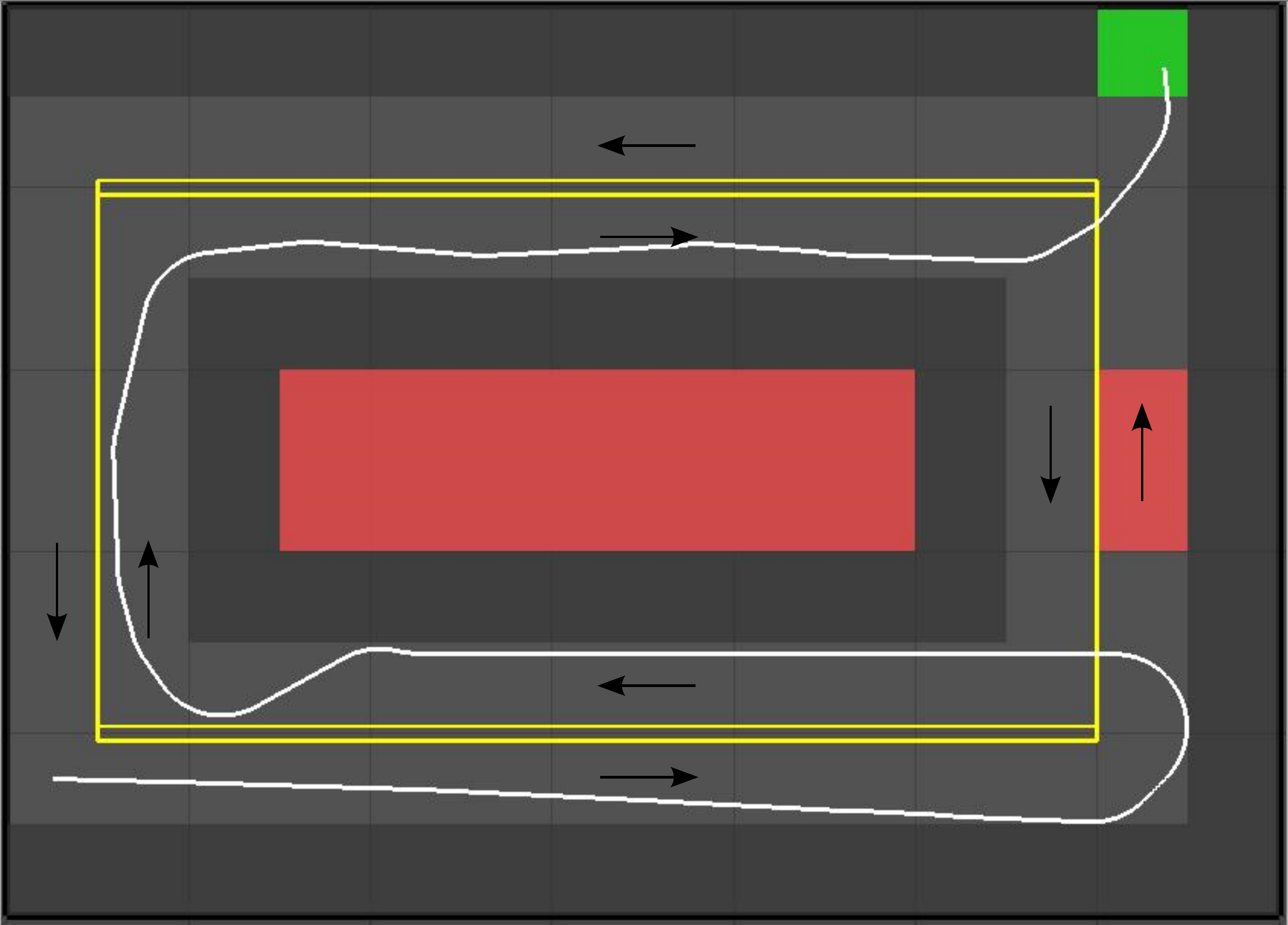}\label{fig:eg2_sw_dir_converge} } 
\\
\subfloat[] {\includegraphics[width=0.22 \textwidth]{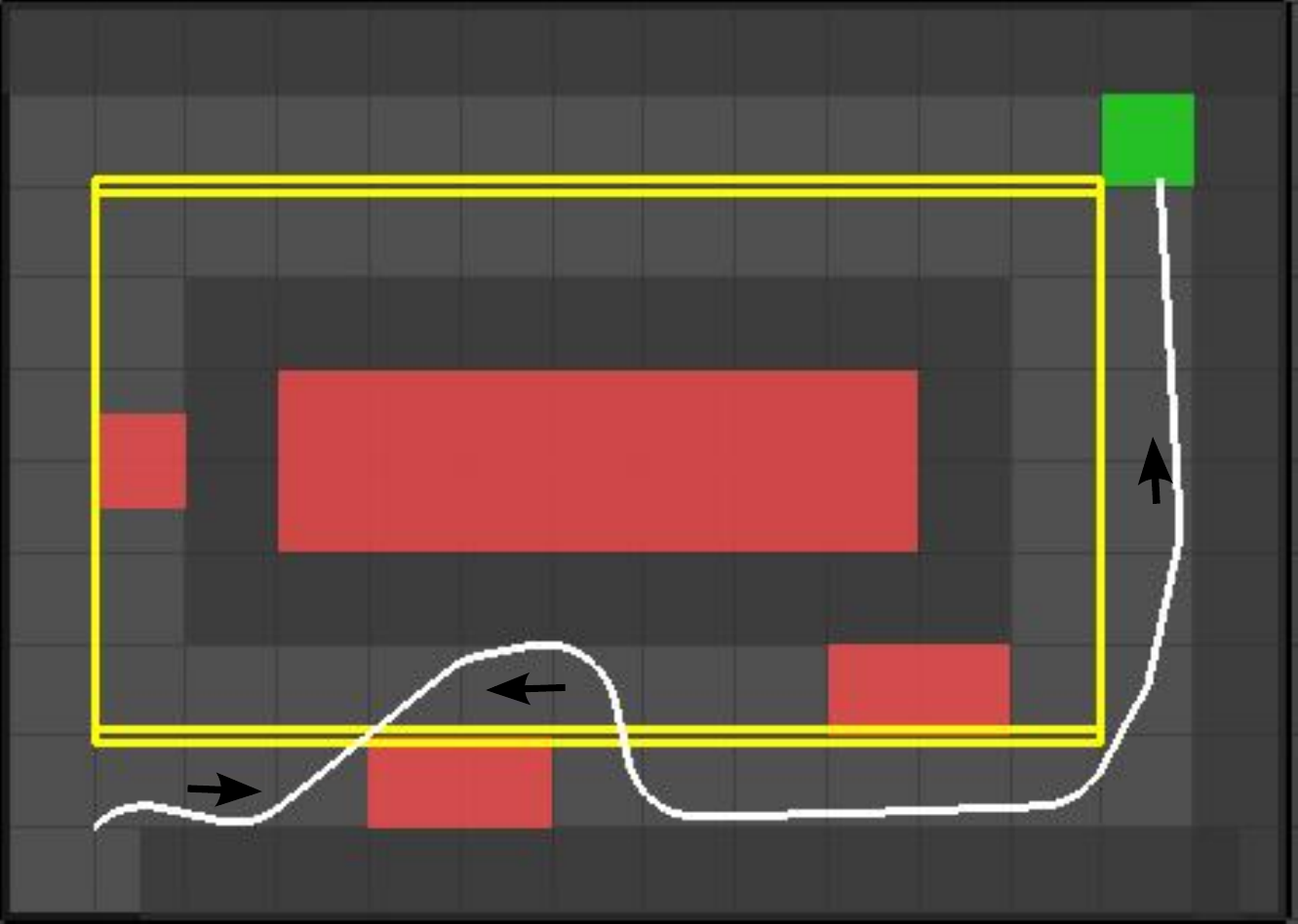}\label{fig:eg2_sw_dir_bold_dotted_no_converge} } \hspace{0.05in}
\subfloat[] {\includegraphics[width=0.22 \textwidth]{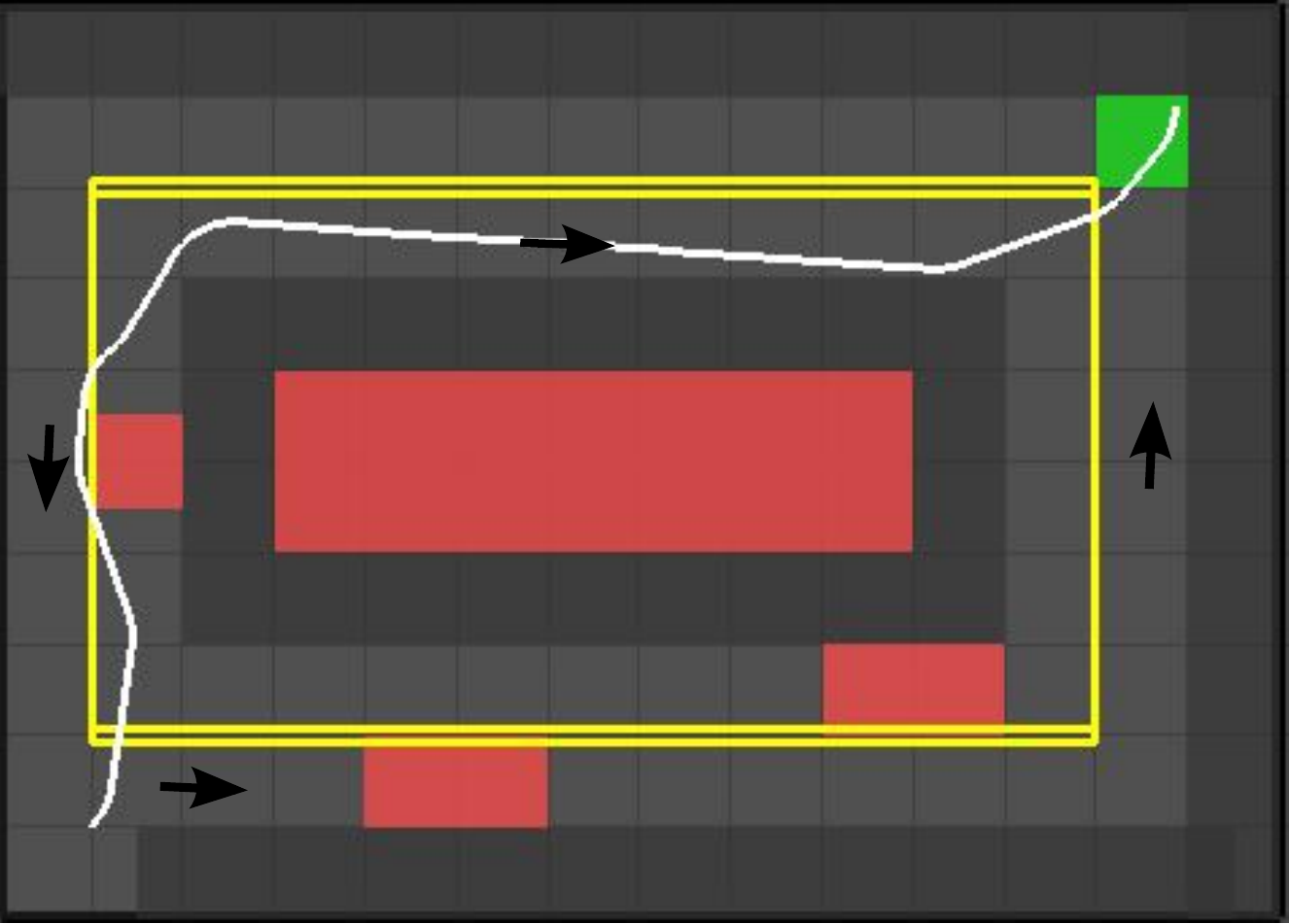}\label{fig:eg2_sw_dir_bold_dotted_converge}}
\caption{
Fig.~\ref{fig:eg2_sw_dir_no_converge} and~\ref{fig:eg2_sw_dir_converge} show the solution of \ALG after 60 and 120 secs. respectively, with the sidewalk, direction and soft lane changing rules. Note that the algorithm converges to a long trajectory which does not break any rules.
Fig.~\ref{fig:eg2_sw_dir_bold_dotted_no_converge} shows a solution after 20 secs. which breaks the hard lane changing rule twice. After 120 secs., the algorithm converges to the solution shown in Fig.~\ref{fig:eg2_sw_dir_bold_dotted_converge}, which features only one hard lane change and three soft lane changes.}
\label{fig:big_eg2}
\end{figure}

\paragraph{Case 3} Fig~\ref{fig:eg2_sw_dir_no_converge} shows a run for the sidewalk, direction and soft lane changing rules after 60 secs. of computation time with a level of unsafety of $(0,0,28.3)$. In Fig.~\ref{fig:eg2_sw_dir_converge}, with 120 secs. of computation, the solution has a much higher cost ($215.8$) but a significantly lower level of unsafety $(0,0,1.6)$ because it only breaks the direction rule slightly when it turns into the lane. This thus demonstrates the incrementality and anytime nature of the algorithm.
\paragraph{Case 4} In our last example, we introduce hard and soft lane changing rules along with sidewalk and direction rules. After 15 secs., \ALG returns the solution shown in Fig.~\ref{fig:eg2_sw_dir_bold_dotted_no_converge}, which breaks the hard lane changing rule twice, thereby incuring a level of unsafety of $(0, 2, 48.1)$ for the three rules. On the other hand, after about 300 secs., the solution converges to the trajectory shown in Fig.~\ref{fig:eg2_sw_dir_bold_dotted_converge} which breaks the hard lane changing rule only once, this has a level of unsafety of $(0, 1, 25.17)$.

\subsection{Implementation}
\label{ssec:implementation}

\begin{figure*}
\centering
\subfloat[] {\includegraphics[width=0.3 \textwidth]{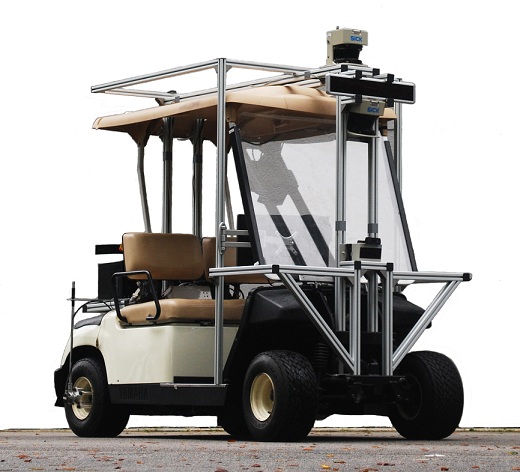} \label{fig:golfcart}} \hspace{0.2in}
\subfloat[] {\includegraphics[scale=0.275]{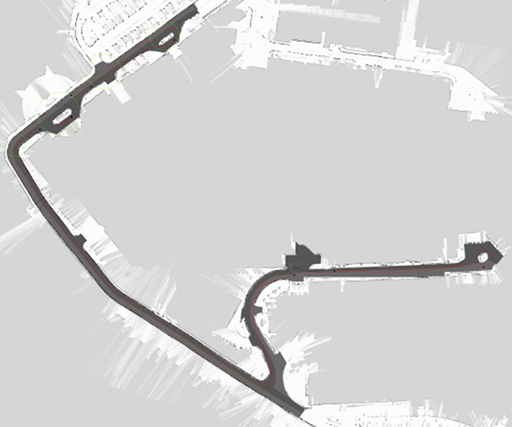} \label{fig:nus}} \hspace{0.2in}
\subfloat[] {\includegraphics[width=0.29 \textwidth]{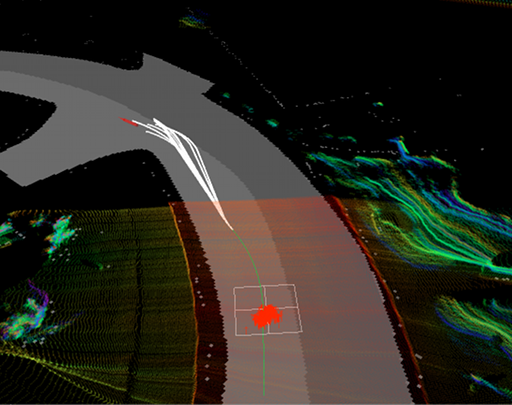} \label{fig:ros}}
\caption{Fig.~\ref{fig:golfcart} shows the Yamaha golfcart instrumented with laser range-finders and cameras. Fig.~\ref{fig:ros} shows the online implementation of \ALG in ROS. Red particles depict the estimate of the current position of the golfcart using laser data (shown using colored points) and adaptive Monte-Carlo localization on a map of a part of the NUS campus shown in Fig.~\ref{fig:nus}. Trajectories of the dynamical system, that are a part of the Kripke structure are shown in white while the trajectory currently being tracked is shown in green.}
\label{fig:golfcart_nus_ros}
\vspace{-0.15in}
\end{figure*}

In this section, we present results of our implementation of \ALG on an autonomous golfcart shown in Fig.~\ref{fig:golfcart_nus_ros} as a part of an urban mobility-on-demand system in the National University of Singapore's campus. The golfcart was instrumented with two SICK LMS200 laser range finders and has drive-by-wire capability. The algorithm was implemented inside the Robot Operating System (ROS)~\cite{quigley2009ros} framework.

\begin{figure}
\centering
\subfloat[] {\includegraphics[width=0.47 \columnwidth] {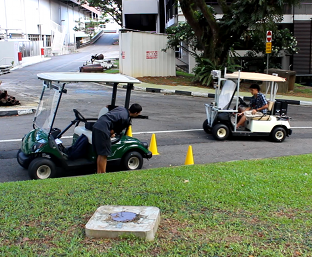} \label{fig:expt_a_1}} \hspace{0.05in}
\subfloat[] {\includegraphics[width=0.47 \columnwidth] {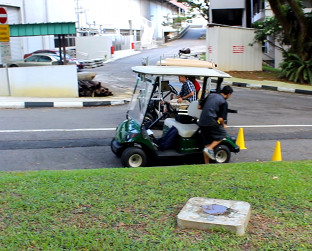} \label{fig:expt_a_2}}\\
\subfloat[] {\includegraphics[width=0.47 \columnwidth] {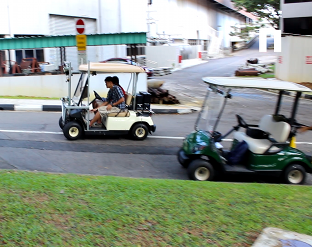} \label{fig:expt_a_3}} \hspace{0.05in}
\subfloat[] {\includegraphics[width=0.47 \columnwidth] {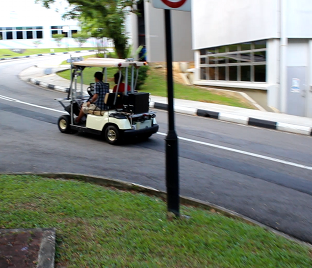} \label{fig:expt_a_4}}
\caption{The autonomous golfcart comes back into the correct lane after overtaking a stalled vehicle inspite of the road curving to the right. Note that the optimal trajectory without road-safety rules would cut through the incoming lane to reach the goal region.}
\label{fig:expt_a}
\end{figure}

Let us briefly describe the setup and note some major implementation details. Traffic lanes and sidewalk regions are detected using pre-generated lane-maps of the campus roads, while obstacles are detected using data from laser range-finders. We use the sidewalk, direction and soft-lane changing rules for the experiments here. For an online implementation of MVRRT$^*$, we incrementally prune parts of Kripke structure that are unreachable from the current state of the golfcart. The algorithm adds new states in every iteration (Lines 5-10 in Alg.~\ref{alg:inc_product}) until the change in the level of unsafety of the best trajectory is within acceptable bounds between successive iterations. This trajectory is then passed to the controller that can track Dubins curves. We use techniques such as branch-and-bound and biased sampling to enable a fast real-time implementation and the golfcart can travel at a speed of approximately 10 kmph while executing the algorithm.
Fig.~\ref{fig:golfcart_nus_ros} gives a snapshot of the experimental setup while Fig.~\ref{fig:expt_a} shows an instance of the golfcart going into the incoming lane in order to overtake a stalled car in its lane. Note that traffic in Singapore drives on the left hand side of the road.


\section{Conclusions}
\label{sec:conclusions}

This paper considered the problem of synthesizing minimum-violation control strategies for continuous dynamical systems that obey a set of safety rules and satisfy a given reachability task. We focused on the case when the task is infeasible without breaking some of the safety rules. Ideas from sampling-based motion-planning algorithms and automata-based model checking approaches were utilized to propose an incremental algorithm to generate a trajectory of the dynamical system that systematically picks which safety rules to violate and minimizes the level of unsafety.
\pcmargin{The algorithm was demonstrated in simulation experiments and also implemented on an experimental autonomous vehicle.}{}

\section{Ackowledgements}
This work is supported in part by Michigan/AFRL Collaborative Center on Control Sciences AFOSR grant FA 8650-07-2-3744, US-NSF grant CNS-1016213, NSF-Singapore through FM SMART IRG, Nissan Motor Company and by the grant LH11065 at Masaryk University, Czech Republic.

\bibliographystyle{unsrt}
\bibliography{reyes.chaudhari.ea.cdc13}

\end{document}